\documentclass[bst/sn-mathphys]{sn-jnl}

\jyear{2021}%

\theoremstyle{thmstyleone}%
\theoremstyle{thmstyletwo}%

\theoremstyle{thmstylethree}%

\usepackage{amssymb}
\usepackage{CJKutf8}
\usepackage{graphicx}

\usepackage{microtype}
\usepackage{adjustbox}
\usepackage{booktabs}
\usepackage{multirow}
\usepackage{subcaption}
\usepackage{color}
\usepackage[marginal]{footmisc}
\usepackage {amsmath}
\usepackage[numbers,sort&compress]{natbib}
\usepackage[misc]{ifsym}

\usepackage{url}

\usepackage{lineno}

\usepackage[english]{babel} 
\usepackage{hyperref}
\hypersetup{
    colorlinks=true,
    linkcolor=red,
    filecolor=red,      
    urlcolor=red,
}

\begin{document}

\title[A Knowledge-enhanced Two-stage Generative Framework for $\ldots$]{A Knowledge-enhanced Two-stage Generative Framework for Medical Dialogue Information Extraction}


\author[1,2]{\fnm{Zefa} \sur{Hu}}
\equalcont{These authors contributed equally to this work.}

\author[1,2]{\fnm{Ziyi} \sur{Ni}}
\equalcont{These authors contributed equally to this work.}

\author[2]{\fnm{Jing} \sur{Shi}}

\author[2]{\fnm{Shuang} \sur{Xu}}

\author[1,2]{\fnm{Bo} \sur{Xu}}

\affil[1]{\orgdiv{School of Artificial Intelligence}, \orgname{University of Chinese Academy of Sciences}, \orgaddress{\city{Beijing} \postcode{100049},  \country{China}}}

\affil[2]{\orgdiv{Institute of Automation}, \orgname{Chinese Academy of Sciences}, \orgaddress{\city{Beijing} \postcode{100190},  \country{China}}}

\abstract{
This paper focuses on term-status pair extraction from medical dialogues (MD-TSPE), which is essential in diagnosis dialogue systems and the automatic scribe of electronic medical records (EMRs). 
In the past few years, works on MD-TSPE have attracted increasing research attention, especially after the remarkable progress made by generative methods. 
However, these generative methods output a whole sequence consisting of term-status pairs in one stage and ignore integrating prior knowledge, which demands a deeper understanding to model the relationship between terms and infer the status of each term. 
This paper presents a knowledge-enhanced two-stage generative framework (KTGF) to address the above challenges. 
Using task-specific prompts, we employ a single model to complete the MD-TSPE through two phases in a unified generative form: 
we generate all terms the first and then generate the status of each generated term. 
In this way, the relationship between terms can be learned more effectively from the sequence containing only terms in the first phase, and our designed knowledge-enhanced prompt in the second phase can leverage the category and status candidates of the generated term for status generation. 
Furthermore, our proposed special status ``not mentioned" makes more terms available and enriches the training data in the second phase, which is critical in the low-resource setting. 
The experiments on the Chunyu and CMDD datasets 
show that the proposed method achieves superior results compared to the state-of-the-art models in the full training and low-resource settings. 
}


\keywords{Medical dialogue understanding, information extraction, text generation, knowledge-enhanced prompt, low-resource setting, data augmentation.
\\
\\
\textbf{Citation:} Z. Hu, Z. Ni, J. Shi, S. Xu, B. Xu. A knowledge-enhanced two-stage generative framework for medical dialogue information
extraction. Machine Intelligence Research, vol.21, no.1, pp.153–168, 2024. \url{http://doi.org/10.1007/s11633-023-1461-5}.
\\
\\
\textbf{Published version:} \url{https://link.springer.com/journal/11633}.
\\
\\
\textbf{Email:} \{huzefa2018, niziyi2021, shijing2014, shuang.xu, xubo\}@ia.ac.cn
}

\maketitle

\section{Introduction}\label{sec1} 
Extracting terms and their statuses from medical dialogues has received increasing attention in the past few years~\cite{2018Annotation,2018An,2018From, quiroz2019challenges, 2020Understanding, liu2020meddg}. 
The extracted information is beneficial to automatic electronic medical records (EMRs) generation~\cite{2020MIE}, which reduces the burden of doctors to document EMRs~\cite{Sinsky2016Allocation,wachter2018combat,2017Tethered}. 
In addition, information extraction is a backbone module of a typical diagnosis dialogue system, where the diagnosis is inferred from the dialogue history~\cite{wei2018task,kao2018context,xu2019end,peng2018refuel}.

Notably, conversation-style medical data in complex speech patterns is challenging because various medical information, such as symptoms, surgeries, and tests, are scattered in a whole turn, even multiple turns.
In addition, colloquial expressions of terms in medical dialogues vary from formal expressions, which further increases the task's difficulty. 
As shown in Table~\ref{example},  the expression of \texttt{\small{Chest pain}} is scattered in the overall sentence, and \texttt{\small{Dyspnea}} is mentioned in the dialogue through the synonymous phrase \texttt{\small{short of breath}}. 
Moreover, the status of each term may be changed as the conversation progresses. 
The word \texttt{\small{No}} indicates the \texttt{\small{absent}} of both \texttt{\small{cardiopalmus}} and \texttt{\small{dyspnea}}, which requires a good understanding of the dialogue. 
The status of \texttt{\small{thyroid function test}} changes from \texttt{\small{suggest}} to \texttt{\small{done}} according to the last turn.

Facing the above challenges, many works~\cite{2019Enhancing,2020MIE, du2019learning, conf/acl/DuCKTCS19, ye2021contrastive, li2021medical, xia2022speaker} have been proposed, 
which can be generally grouped into two categories: 
classification-based methods and generative methods. 
One way~\cite{2020MIE, xia2022speaker} of the classification-based methods takes each term candidate as the input to model the semantic interaction between the medical dialogue and the candidate, then determines the status of the term candidate by classification. 
However, this way treats each term independently, ignoring their relationship.
Another way~\cite{2019Enhancing, du2019learning, conf/acl/DuCKTCS19} is to decompose the task into multiple stages, including term detection, term normalization, and status inference. 
It first detects colloquial expressions of terms in dialogue, then maps the colloquial expressions to formal expressions and infers the status. 
However, term detection in the first stage requires token-level annotation. 
The generative methods~\cite{conf/acl/DuCKTCS19, ye2021contrastive, li2021medical} cast the task as a sequence generation problem regarding formal expressions of terms and status candidates as a target vocabulary and generating terms and the corresponding status sequentially. 
In this way, generative methods need not explicitly detect terms. 
Consequently, it requires only the final formal label rather than token-level BIO (begin-in-out) annotations. 
At the same time, sequence generation can model the relationship between the terms.

\begin{table}[t]
  \centering
  \caption{\label{example}An example of term-status pair extraction from medical dialogues. 
}
    \scalebox{0.85}{
    \begin{tabular}{ll}
    \toprule
    \multicolumn{2}{l}{\textbf{Dialogue}} \\
    \midrule
    Patient: & \multicolumn{1}{p{21.445em}}{\begin{CJK*}{UTF8}{gbsn}我有房颤，怎么治疗？\end{CJK*}} \\
          & \multicolumn{1}{p{21.445em}}{I had atrial fibrillation, how to treat it?} \\
    Doctor: & \begin{CJK*}{UTF8}{gbsn}有没有心慌、气短？\end{CJK*} \\
          & Do you feel palpitation or short of breath? \\
    Patient: & \begin{CJK*}{UTF8}{gbsn}没有，但胸口会不舒服，经常疼\end{CJK*} \\
          & No, but my chest is uncomfortable, always has bouts of pain \\
    Doctor: & \begin{CJK*}{UTF8}{gbsn}先完善一下甲状腺功能检查，正常的话，建议射频消融治疗\end{CJK*} \\
          & \multicolumn{1}{p{21.445em}}{First, complete the thyroid function test. If it is normal, radiofrequency ablation is recommended.} \\
    Patient: & \begin{CJK*}{UTF8}{gbsn}检查已经做过了。\end{CJK*} \\
          & The examination has been done. \\
    \midrule
    \multicolumn{2}{l}{\textbf{Term-Status Pairs Label}} \\
    \midrule
    \multicolumn{2}{l}{\begin{CJK*}{UTF8}{gbsn}房颤: 阳性\end{CJK*} (Atrial fibrillation: appear)} \\
    \multicolumn{2}{l}{\begin{CJK*}{UTF8}{gbsn}心慌: 阴性\end{CJK*} (Cardiopalmus: absent)} \\
    \multicolumn{2}{l}{\begin{CJK*}{UTF8}{gbsn}呼吸困难: 阴性\end{CJK*} (Dyspnea: absent)} \\
    \multicolumn{2}{l}{\begin{CJK*}{UTF8}{gbsn}胸痛: 阳性\end{CJK*} (Chest pain: appear)} \\
    \multicolumn{2}{l}{\begin{CJK*}{UTF8}{gbsn}射频消融: 医生建议\end{CJK*} (Radiofrequency ablation: suggest)} \\
    \multicolumn{2}{l}{\begin{CJK*}{UTF8}{gbsn}甲状腺功能: 患者已做\end{CJK*} (Thyroid function test: done)} \\
    \bottomrule
    \end{tabular}}%
\end{table}%

Although existing generative methods have many advantages compared with other methods, 
there are still some limitations: (1) Complex relationship modelling among terms. 
The output sequence of these generative methods consists of terms and status. 
If the model explores the association between terms through this sequence, it must understand the corresponding status simultaneously.
(2) Prior schema information is neglected. These generative methods output the sequence based only on the dialogue history. The schema information for status generation can only be learned through mapping from medical dialogue to the output sequence. 
In fact, 
some status candidates are predefined in medical datasets that could guide the model to generate status but are ignored. 
(3) Lack of training data in the low-resource setting. The training data of these generative methods must be fully annotated (including terms and status). These methods preclude the model from using the existing data that only has the term annotation, which is unfriendly to the low-resource setting.

\begin{figure*}[t]
\centering{ 
\includegraphics[height=6cm,width=12cm]{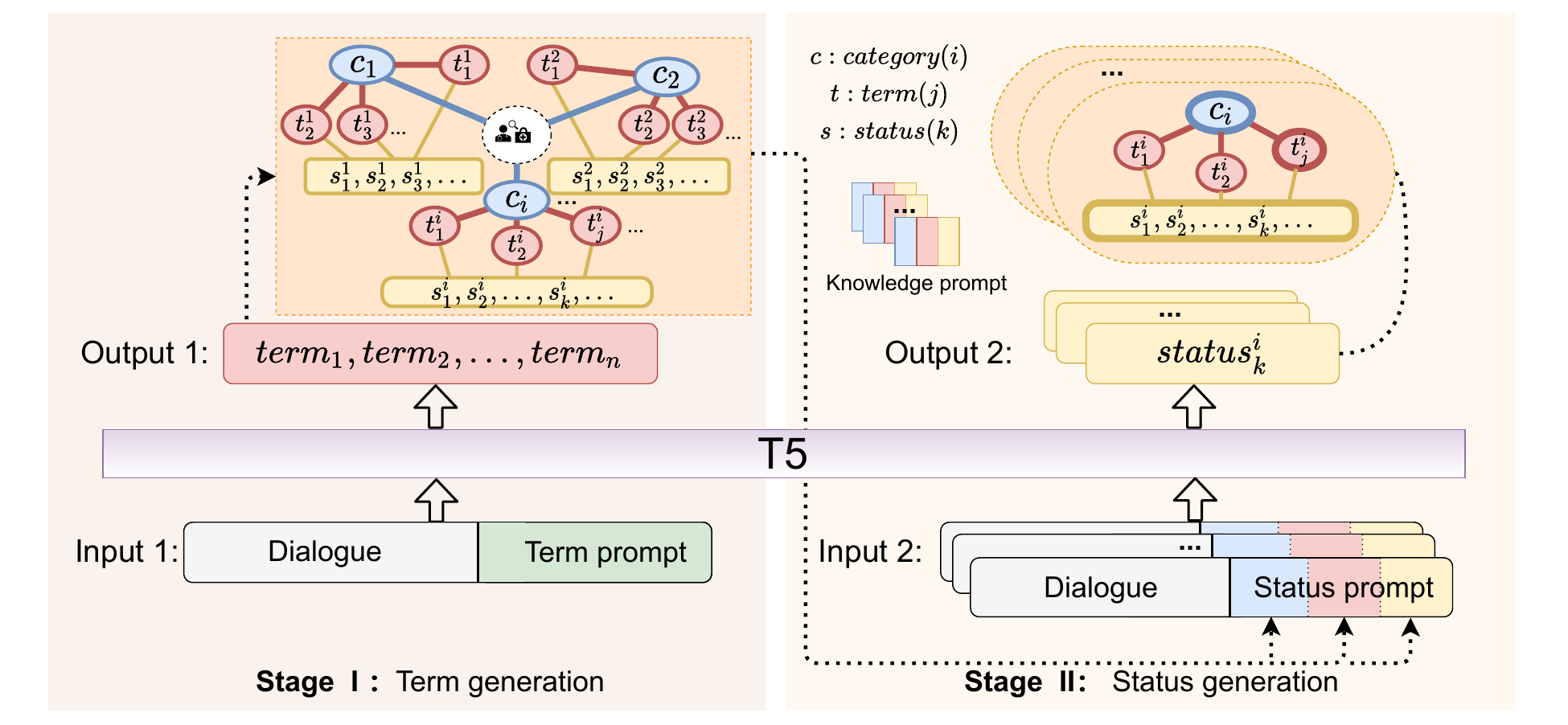}}
\caption{
Overview of KTGF. 
$t^{i}_{j}$ and $s^{i}_{k}$ indicate the $j$-th term and the $k$-th status in the $i$-th category, respectively. 
KTGF takes T5 as the backbone and generates terms and their status in two stages.
In each stage, KTGF concatenates medical dialogue with a subtask prompt as input. 
In the second stage, KTGF uses the terms generated in the first stage to retrieve prior task knowledge to enhance the prompt, which enables the model to generate status effectively.
}
\label{Framework}                     
\end{figure*}

In this paper, we propose a novel knowledge-enhanced two-stage generative framework (KTGF). 
The overview of KTGF is shown in Figure \ref{Framework}. 
In our framework, we complete term-status pair extraction from medical dialogues (MD-TSPE) through two phases in a unified form of the sequence 
to sequence generation: 
we generate all terms first and then generate the status of each generated term. 
We also plug different task-specific prompts after the dialogue context and take them together as the model input. 
In this way, the generation of terms and their status are decoupled, leading to greater flexibility of the framework that introduces at least three advantages: 
(1) The sequence contains only terms in the first phase, which enables the model to learn the relationship between terms more effectively without considering the status. 
(2) The prior schema information can be fully utilized. 
We integrate the status candidates and the category of generated terms defined in the schema into the prompt of the second phase and name it the knowledge-enhanced prompt. 
It allows task-aware contextualization and can guide the model to generate status more effectively. 
(3) As terms and their statuses are generated separately, the framework can learn from data that only has term annotations in the first phase, which can partially alleviate the data sparsity problem. 
In addition, we design another status, ``not mentioned", to further enrich the data and enhance the training of the second phase.
We evaluate KTGF on two medical dialogue datasets, Chunyu~\cite{2020MIE} and CMDD~\cite{2019Enhancing}. Comparisons against previous state-of-the-art methods show that KTGF performs best in both the full training and the low-resource setting.

Our main contributions can be summarized as follows\footnote{Our codes are available at https://github.com/FlyingCat-fa/KTGF.}: 
\begin{itemize}
\item We propose a novel knowledge-enhanced two-stage generative framework (KTGF) to explore the term-status pair extraction from medical dialogues(MD-TSPE). Our framework only needs to consider the relationship between terms in the first phase. Then it generates the corresponding status from the dialogue in the second phase with the help of prior schema knowledge in the prompt. 
\item KTGF can utilize data with only term annotation, which is critical in the low-resource setting. The extra-designed status further enriches the status-related data to alleviate the data sparsity problem. 
\item We evaluate KTGF on Chunyu~\cite{2020MIE} and CMDD~\cite{2019Enhancing}. 
Our approach achieves new state-of-the-art results on both datasets. 
Ablation studies and further analysis demonstrate the advantages of our two-stage generation strategy and prior knowledge utilization.
\end{itemize}

The remainder of this paper is organized as follows. 
Section \ref{sec:Related Work} presents previous works related to 
medical information extraction, prompting language models, and training on supplementary data. 
Section \ref{sec:Methodology} introduces the problem definition and 
proposes our KTGF. 
Section \ref{sec:Experiments} describes the datasets, compared baselines, experimental settings, and results. 
In Section \ref{sec:Analysis}, further analysis and a discussion are presented. 
Finally, the conclusion is shown in Section \ref{sec:Conclusion}.

\section{Related Work}
\label{sec:Related Work}

\subsection{Medical Information Extraction}
Medical Information Extraction has great potential in the fields of biomedicine and NLP. 
One of the classic tasks is the \emph{i2b2 challenge}. 
It constructs a small corpus of written discharge summaries 
and designs a variety of tasks to extract different kinds of information~\cite{uzuner20112010}. 
Many efforts have also focused on extracting entities and relations from medical texts\cite{lai2021joint, su2022deep}, which is also taken as one of the tasks for Chinese Biomedical Language Understanding Evaluation (CBLUE)\cite{zhang2022cblue}. 
In contrast, extracting information directly from medical dialogues has been an emerging research field in recent years~\cite{happe2003automatic, quiroz2019challenges, hu2023matching}, focusing on colloquialism and multi-turn dialogue interaction. 
Given the patient-doctor conversation, medical information, including clinical terms~\cite{2020Understanding,shi2021understanding} and their properties~\cite{conf/acl/DuCKTCS19, du2019learning}, or treatment regimens~\cite{yan2020knowledge} can be extracted and used to generate electronic medical records (EMRs)~\cite{2018An, finley2018dictations}, reducing the burden on doctors of creating narrative reports~\cite{Sinsky2016Allocation,wachter2018combat,2017Tethered}. 
Information extraction from medical dialogues is also a backbone module of the typical diagnosis system~\cite{wei2018task,kao2018context,xu2019end,peng2018refuel}, in which the status information of medical terms plays a key role.

Among the works on medical dialogue information extraction,
~\cite{2018An} extracts information from dialogue through a pipeline 
consisting of various heuristics such as character matching, regular expressions, and other task-specific heuristics. 
~\cite{finley2018dictations} proposed generating EMRs from an automatic speech recognition (ASR) directly, but it achieved poor performance. 
Recently, classification-based methods have been proposed, including multi-stage methods~\cite{conf/acl/DuCKTCS19, du2019learning, 2019Enhancing} and matching-based classification methods~\cite{2020MIE, xia2022speaker}. 
The multi-stage methods identify the term by a sequence tagging model in the first stage 
and then conduct term normalization and status inference by two multi-layer perception classifiers. 
However, these methods require token-level redundant annotation in the first stage. 
In the matching-based classification methods, 
~\cite{2020MIE} takes each term, corresponding category, and status candidates as input to match the semantics with the medical dialogue, and then designs an aggregate module to consider the utterance interaction. 
SAFE~\cite{xia2022speaker} also takes each term candidate as input. The method develops a multi-task learning method to model the speaker's identity. It further proposes a co-attention fusion module with graph networks to match the semantics between the medical dialogue and the term. 
However, these matching-based classification methods do not consider the relationship between terms. 
The generative method is considered an excellent way to address the problems in classification methods. 
The generative methods~\cite{conf/acl/DuCKTCS19, li2021medical, ye2021contrastive} regard formal medical terms and status as the target vocabulary and generate a single sequence containing terms and their status, which gain better performance compared with classification-based methods.
Based on these generative methods, we aim to further enhance the modeling of relationships among terms and fully use prior schema knowledge.

\subsection{Prompting Language Models}
Combining the prompt with pre-trained language models is an active line of research~\cite{liu2023pre}. 
~\cite{radford2019language} uses prompts for generation in the zero-shot setting and obtains excellent performance. 
~\cite{raffel2020exploring} designed task-specific prompts for training and testing to learn multiple tasks simultaneously. 
Many works also focus on automatically optimizing discrete prompts~\cite{gao2020making, jiang2020can} or directly using learnable continuous prompts~\cite{lester2021power, li2021prefix}. 
In addition, to prompt engineering, some works also use available prior task information or knowledge as task-specific prompts~\cite{lee2021dialogue, hu2022knowledgeable, han2022ptr, chen2022knowprompt}.
Our proposed prompting method inspired by these considerations considers both multi-task and knowledge prompts. 
Moreover, we further improve the prompt to enrich training data in the low-resource setting. 

\subsection{Training on Supplementary Data}
Previous works have shown that the performance of the target task can be improved through supplementary training on tasks with intermediate-labelled data~\cite{phang2018sentence, aghajanyan2021muppet, su2022multi}.
~\cite{phang2018sentence} and~\cite{aghajanyan2021muppet} focus on GLUE natural language understanding benchmark~\cite{wang2018glue}, 
and~\cite{su2022multi} improves the task-oriented dialogue system through an augmented training phase with intermediate-labelled data.
In this work, we aim to boost the performance of term-status pair extraction from medical dialogues through more supplementary data, including not only the intermediate-labelled data, but also the additional data available owing to our designed special status.

\section{Methodology}
\label{sec:Methodology}
In this section, we will define the term-status pair extraction from medical dialogues (MD-TSPE) task and then introduce our approach, including an overview of the approach, the two-stage generative framework, prompt design with prior schema knowledge, and training strategies.

\subsection{Problem Definition}
\label{sec:Problem Definition}

\textbf{Medical Dialogue} $D$: The medical dialogue is the raw text containing multiple sentences alternated between the patient and the doctor, namely $D=(l_1, l_2, \ldots, l_n)$, 
which consists of a sequence of word tokens $(w_1, w_2, \ldots, w_m)$.

\textbf{Term} $T$: The term set $T$ is predefined according to medical knowledge graphs. 
Each term $t_i$ can be mentioned in the medical dialogues in different ways, 
such as formal expression, spoken word expression, 
and the expression scattered in the whole sentence or even across multiple sentences. 
Each term belongs to a predefined category in the dataset. 

\textbf{Status} $S$: The status is essential because it specifies more detailed information about the term, 
and status candidates depend on each term's category. 
Furthermore, the status of each term is changeable during the whole medical dialogue.

Given Medical Dialogue $D$, 
the objective of MD-TSPE
is to extract 
a set of term-status pairs $\{\ldots, (t_i, s_i), \ldots\}$, 
where $t_i \in T$ is the term mentioned in the dialogue, and 
$s_i \in S$ is the corresponding status of $t_i$.

\subsection{Overview of Our Approach}
Previous generative methods on MD-TSPE cast the task as a sequence generation problem in one stage.
In this case, the medial dialogue $D$ is taken as input to the model, 
and term-status pairs $\{\ldots, (t_i, s_i), \ldots\}$ are generated sequentially. 
This approach has been adopted in many works, such as LSTM-based or transformer-based models~\cite{conf/acl/DuCKTCS19, li2021medical}. 
However, generating a single output sequence consisting of both terms and their status is more likely to hardly model the relationship among terms and suffers optimization issues with decoding long and complex sequences, which results in poor performance. 
~\cite{ye2021contrastive} leverages contrastive learning to filter out the unfaithful term-status pairs after generating complex and long output sequences. 
However, this method also needs to generate long and complex sequences, resulting in missing some term-status pairs. 
Even so, considering the common use of the one-stage generation approach, we first include this basic framework as our preliminary experiments.

Then, we design a knowledge-enhanced two-stage generative framework (KTGF) for MD-TSPE, of which each stage has its corresponding task. 
In the first stage, we aim to generate a simplified sequence containing only the terms to model the relationship among them and then generate the corresponding status of each generated term in the second stage. 
Thus, the complex sequence in the one stage generation approach is disintegrated into multiple sequences.
To make the form of our framework unified in two stages, 
we plug different task-specific prompts after the dialogue context and take them together as the model input. 
In addition, prior knowledge can enrich the task-specific prompt in the status generation phase and further benefit generative methods. 
Moreover, we design a special status that improves the status prompt to expand the status-related training data in the low-resource setting.

\subsection{A Two-stage Generative Framework for MD-TSPE}

Given the medical dialogue, we concatenate each utterance into a word token sequence $ D = (w_1, w_2, \ldots, w_m)$. 
The basic idea is to encode the medical dialogue $D$, obtain its contextualized representations, and generate the medical terms and their status in two stages. 
Notably, the generation in two stages can be fulfilled in a unified generative model. 
In the second stage, each generated term in the first stage needs to be appended to medical dialogue $D$. 
However, the generation of two stages is actually different subtasks. 
To address this issue, we adopt the prompting approach, where elaborately designed prompt tokens are concatenated with the original medical dialogue. The knowledge of generated terms is incorporated into the prompt of the second stage. 
It has been confirmed that prompting is an adequate paradigm to leverage the prior information for specific tasks~\cite{lee2021dialogue, hu2022knowledgeable}, or the knowledge of pre-trained language models (PLMs) to solve various tasks~\cite{brown2020language, gao2020making}.
After incorporating the prompts, the original medical dialogue $D$ can be extended to a task-specific sequence, denoted as $\widetilde{D}$:
\begin{equation}
    \widetilde{D} = \underbrace{w_{1}, \cdots, w_{m},}_{\text{word tokens}} \underbrace{p_1,\dots, p_{n}}_{\text{prompt tokens}}.
\end{equation}

Given the task-specific sequence $\widetilde{D}$ as the input, we add a special token $[SOS]$ to represent the beginning of each target sequence. 
Furthermore, a special end-of-sequence $[SEP]$ is also appended to the end of the target sequence. 
The target sequence of the first stage consists of all terms mentioned in the medical dialogue $D$, and we use a special token `` , " to separate different terms. 
The status of the generated term that can be found from the prompt is the target sequence of the second stage.

We employ an encoder-decoder transformer architecture T5~\cite{raffel2020exploring} as the backbone of our framework. 
This framework first embeds the input 
sequence as the input vector representation: 
\begin{equation}
    H^{0} = Emb(S),
\end{equation}
where $S$ denotes the embedded sequence.
Based on the input representation $H^{0}$, both the transformer encoder and decoder employ $L$ stacked transformer blocks, and $l$-th block learns a new sequence representation $H^{l}= H^{l}_{1}, \cdots, H^{l}_{n}$ from $H^{l-1}= H^{l-1}_{1}, \cdots, H^{l-1}_{n}$. 
Each layer of the encoder has two sub-layers. 
The first is a multi-head self-attention mechanism, and the second is a small position-wise fully connected feed-forward network. 
Layer normalization~\cite{ba2016layer} is applied to the input of each sub-layer, and the applied layer normalization is a simplified version where no additive bias is utilized. 
Following layer normalization, a residual skip connection adds each sub-layer's input to its output. 
In addition to the applied two sub-layers in each encoder layer, the decoder layer inserts a third sub-layer, which has the same structure as the first sub-layer and performs attention over the output of the encoder. 
The following are the details of each sub-layer.

The first sub-layer is multi-head self-attention, which computes independent self-attention representations with multiple individual heads. 
For each head,
the input consists of queries $Q$ and keys $K$ of dimension $d_{k}$, values $V$ of dimension $d_{v}$. 
The output is a weighted sum of values, and the weights are calculated by the scaled dot-product between queries $Q$ and keys $K$. 
Since the transformer mainly relies on the attention mechanism without recurrence, 
the transformer adds a scalar as the position embedding parameter to the corresponding logit for computing the attention weights. 
The learnable position embedding is assigned based on the relative positions between queries $Q$ and keys $K$ and shared across all layers in the model.
The calculation of a self-attention head is shown as follows:
\begin{equation}
    \textbf{$\operatorname{Attention}(Q,K,V)=\operatorname{softmax}\left(\frac{\mathbf{Q} \mathbf{K}^{\top}}{\sqrt{d_{k}}}+\mathbf{M}+\mathbf{PE}\right)\mathbf{V}$},
\end{equation}
\begin{equation}
\mathbf{M}_{i j}=\left\{\begin{array}{ll}0, & \text { visible } \\ -\infty, & \text { invisible }\end{array}\right.
\end{equation}
where $d_{k}$ is the dimension of the key. 
$\mathbf{M}, \mathbf{PE} \in \mathbb{R}^{n \times n}$ are an attention mask and the position embedding matrix, and $n$ is the sequence length. 
The attention mask $\mathbf{M}$ can makes a key invisible or no contribution to a query by setting the attention score $-\infty$.

Multi-head attention allows the model to compute multiple subspace representations in parallel and concatenate them.
\begin{equation}
    \mathrm{head_i} = \mathrm{Attention}(H^{l-1}W^Q_i, H^{l-1}W^K_i, H^{l-1}W^V_i),
\end{equation}
\begin{equation}
    \operatorname{MHAttention} = \mathrm{ConCat}(\mathrm{head_1}, ..., \mathrm{head_h})W^O,
\end{equation}
where $H^{l-1}$ and $h$ are the input of the $l$-th block and the number of attention heads, respectively.
$W^Q_i$, $W^K_i$, $W^V_i$ and $W^O$ are learnable parameters.

The second sub-layer is a simple position-wise fully connected feed-forward network. 
Following layer normalization, a residual skip connection adds the sub-layers' input to their output. 
\begin{equation}
    O^l = H^{l-1} + \operatorname{LayerNorm}(\operatorname{MHAttention}(H^{l-1})),
\end{equation}
\begin{equation}
    H^{l} = O^l + \operatorname{LayerNorm}(\operatorname{}FFN(O^l)),
\end{equation}
where FFN is the feed-forward layer:
\begin{equation}
   \mathrm{FFN}(x)=\max(0, xW_1 + b_1) W_2 + b_2,
\end{equation}
where $W_1$, $W_2$, $b_1$ and $b_2$ are trainable model parameters.

Based on the transformer-based encoder-decoder architecture, 
T5 designs an unsupervised objective that randomly samples and then drops out the sampled tokens in the input sequence. 
A single sentinel special token replaces all consecutive spans of dropped-out tokens. 
The target is to recover all corresponding dropped-out spans of tokens delimited by the same sentinel special tokens used in the input sequence. 

Taking the pre-trained T5 as the backbone,
we further train our model to generate the term and status sequences in two stages. 
Each training sample is represented as follows:
\begin{equation}
    d = (\widetilde{D}_t, y), 
\end{equation} 
where $t\in\{\textup{term generation}, \textup{status generation}\}$, which denotes the subtask that sample $d$ belongs to. 
$\widetilde{D}_t$ is the subtask input that consists of the medical dialogue and the subtask prompt. 
$y$ denotes the subtask output text.
We train the model with the maximum likelihood objective. 
Given the training sample $d = (\widetilde{D}_t, y)$, the objective $\mathcal{L}_{\Theta}$ is defined as:
\begin{equation}
    \label{eq:mle}
    \mathcal{L}_{\Theta} = -\sum_{i=1}^{|y|}\log P_{\Theta}(y_{i}\vert y_{<i}; \widetilde{D}_t),
\end{equation}
where $\Theta$ is the model parameters.

\begin{figure*}[t]
\centering                   
\includegraphics[height=4cm,width=12cm]{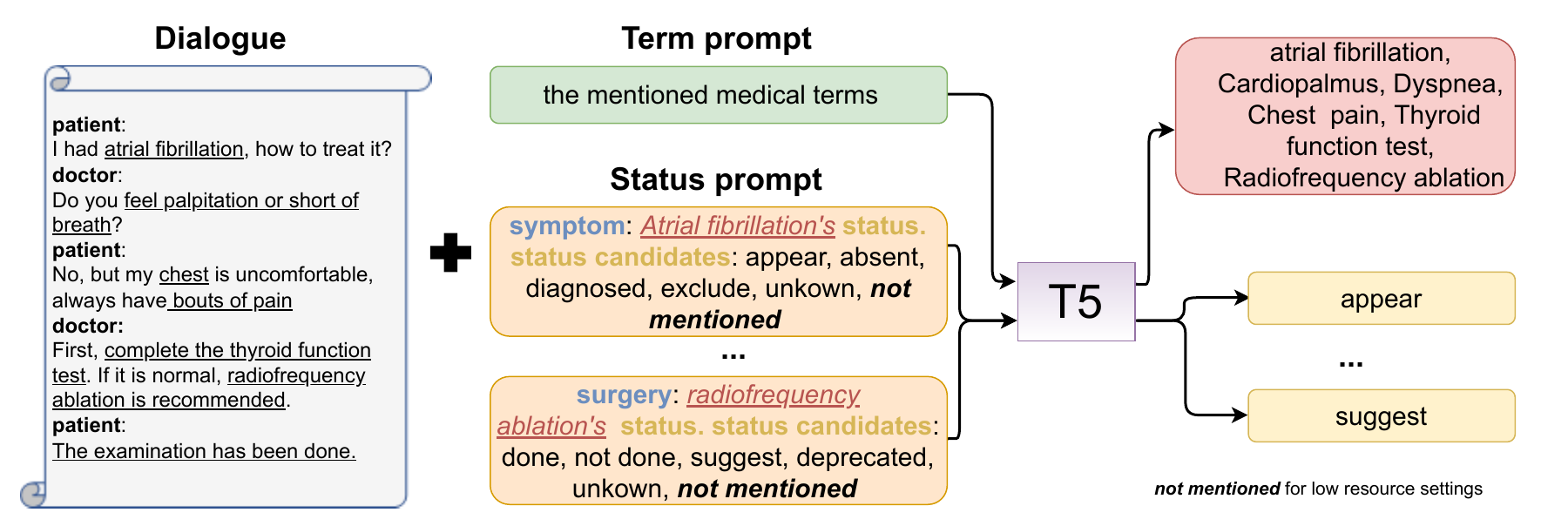}
\caption{The design of our prompt. 
In the term generation stage, the prompt is employed for a better understanding of the term generation subtask. 
In the status generation stage, we use the generated term to obtain the category and status candidates, which are utilized to enhance the prompt. 
Moreover, we add a special status, "not mentioned", for the low-resource setting. 
Therefore, the term not mentioned in the dialogue can also have its corresponding status, which augments the status-related training data. 
The medical dialogue is from Table \ref{example}.
}
\label{prompt}                     
\end{figure*}                    

\subsection{Knowledge-enhanced Prompt}
Our framework uses the same T5 to generate terms and status in two stages, which 
requires the model to learn two different subtasks. 
Therefore, we design task-specific prompts to make better use of the knowledge of the pre-trained T5 and jointly learn the two subtasks.
We further incorporate the prior schema information of each generated term as additional prompt tokens for status generation. 
This paper further considers the low-resource setting.
In reality, data with term and status annotation is usually more difficult to obtain. 
Compared with the existing generative methods, our model has the advantage of being able to learn from other data containing only term annotation in the first stage. 
Moreover, we provide a status-enhancing method that improves the status prompt to expand the status-related training data in the low-resource setting. 
Next, we describe the specific prompting designs of the two stages in detail.

\textbf{Prompt for term generation.} 
Term generation aims to generate all terms mentioned in the medical dialogue. 
The prompting design of term generation mainly enhances the textual semantics for better medical dialogue understanding and term generation. 
Here we set ``the mentioned medical terms" as the prompt and add it after the medical dialogue.

\textbf{Prompt for status generation.} 
\label{sec:status prompt}
After obtaining the generated term, we further generate the corresponding status of each generated term in the status generation stage.
The prompt for status generation is designed to enhance the term semantics and predict status more accurately. 
For each generated term, we retrieve the prior schema information, including the category and status candidates of the term. 
The information is leveraged to construct a knowledge-enhanced status prompt, 
which is formally denoted as: ``\texttt{\small{Category:}} \texttt{\small{term's status.}} 
 \texttt{\small{Status candidates: candidate $1$, candidate $2$, etc ... }}."
The knowledge-enhanced prompt advances task-aware contextualization and guides the model to generate status more effectively.

Moreover, we improve the knowledge-enhanced prompt for status generation in the low-resource setting. 
Specifically, we add a special status, ``not mentioned", to the status candidates. 
In this way, many terms not mentioned in a medical dialogue can match the special status "not mentioned" as a term-status pair.
Therefore, the training data of status generation increases, 
and the model can learn how to generate the correct status according to the status candidates in the low-resource setting. 
The design of term and status prompts is also shown in Figure \ref{prompt}.

\subsection{Training Strategies}
We employ multi-task learning approach to train a single unified model in the two-stage generative framework and use a mini-batch based optimization approach. 
We adopt different prompting strategies for full-data and low-resource settings.
In the full-data setting, we adopt the knowledge-enhanced prompt without additional special status. 
In the low-resource setting, we improve the knowledge-enhanced prompt by introducing a special status as described in Section \ref{sec:status prompt}. 
KTGF is trained in two stages in the low-resource setting. The model is first trained on a mixed dataset consisting of the training set and external corpora containing only term annotations, and then further fine-tuned on the training set for better performance. 
In this way, the partially annotated data containing only terms can also be leveraged in the low-resource setting. 
More experimental details are introduced in Section \ref{sec:Experiments Details}.

\section{Experiments}
\label{sec:Experiments}

\subsection{Datasets and Metrics}
\label{sec:Datasets and Metrics}

\begin{table}[t]
    \centering
    \begin{adjustbox}{scale=0.9,center}
    \begin{tabular}{lccccc}
    \toprule
    Dataset & Domain & Dialogue & Window & Term & Status \\
    \midrule
    Chunyu   & Cardiology & 1,120 & 18212 & 71    & 18 \\
    CMDD  & Pediatrics & 2,067 & 87005 & 149   & 3 \\
    \bottomrule
    \end{tabular}
    \end{adjustbox}
    \caption{\label{dataset} The statistics of Chunyu dataset and CMDD dataset.}
\end{table}

We evaluate our proposed framework on two benchmark datasets: 
Chunyu~\cite{2020MIE} and CMDD~\cite{2019Enhancing}, both of which are Chinese medical dialogue datasets for entity-status pair extraction.

Chunyu contains $1,120$ dialogues on cardiovascular diseases. 
A schema is defined in the dataset, which contains four main categories: symptom, surgery, test, and other info. 
Each category includes medical terms and possible statuses.
The original Chunyu dataset provides two settings for the status, including the coarse-grained status independent of categories and the fine-grained status dependent on categories~\cite{2020MIE}. This work uses the fine-grained status better to express the status semantics of terms from different categories. 
For example, surgery includes the term, radiofrequency ablation, and its status candidates consist of done, not done, suggest, deprecated, and unknown. 
The schema has $71$ predefined terms and $18$ predefined statuses. 
Chunyu is annotated by a window-to-information approach with the mentioned terms and corresponding status, which focuses on 
the dialogue context in the window and does not need word-level sequence labelling information. 
Specifically, the dialogues in the dataset are divided into pieces using a sliding window. 
Each window consists of $5$ turns of the dialogue. 
The sliding step is set to $1$. 
In this way, the dataset obtains $18212$ annotated windows.

CMDD has $2,067$ annotated dialogues on four pediatric diseases, 
which contain $149$ predefined terms and $3$ predefined statuses. 
All of these terms belong to the symptom category, and the status candidates are True, False, and Uncertain. 
Each term-related character in the dialogue turn is annotated by the word-level begin-in-out (BIO) label. 
In the original data, multiple adjacent sentences belonging to the same speaker are considered  multiple turns. 
Following~\cite{li2021medical}, we merged these adjacent sentences from one speaker into a single dialogue turn. 
We also split the medical dialogues into windows and kept the window size and sliding step the same as Chunyu.

We followed~\cite{li2021medical} to further process the status annotation in the two datasets.
In the original annotation of the Chunyu dataset, one window contained several statuses for the same term. 
Compared to the historical status in medical dialogues, the latest status of each term is more informative.
We only preserve the latest status, which is more suitable for the medical dialogue system in the experimental setting~\cite{li2021medical}.
In fact, the historical status of each term in the current window can be obtained through the latest status of the term in the historical window, so there is no historical status information loss.
In the CMDD dataset, the repetitive medical term-status pairs in one window are removed.
Table~\ref{dataset} summarizes the statistics of the Chunyu dataset and CMDD dataset. 

Following~\cite{2020MIE}, we use the micro-average precision, recall, and F1-score as the evaluation metrics. 
We report the results of extracted medical information in term and full evaluations. 
Term evaluation only considers the correctness of the terms, regardless of their status. 
Full evaluation means that both terms and the corresponding status must be strictly correct.
The evaluation results for both window-level and dialogue-level on the test set are reported. In line with~\cite{2020MIE}, we merge the results of windows within the same dialogue for the dialogue-level evaluation and update the previous labels with the latest ones. 
Notably, not all medical windows have golden labels; sometimes, the windows are not associated with medical terms in the dataset schema. 
In this case, the empty prediction is regarded as correct, and then the precision, recall, and F1 are set to $1$; otherwise, they are $0$.

\subsection{Baseline Models}

We compared our proposed method KTGF against classification-based and generative models for MD-TSPE.

\noindent\textbf{SAT}~\cite{conf/acl/DuCKTCS19}: a multi-stage classification-based method. 
It identifies the term spans by a sequence tagging model trained with token-level annotation in the first stage. Then, it employs two multi-layer perception classifiers to normalize the terms and infer their status from the hidden representation of identified term spans.

\noindent\textbf{BERT-SAT}: a multi-stage classification-based method similar to SAT, except that the encoder is replaced by BERT~\cite{devlin-etal-2019-bert}.

\noindent\textbf{MIE}~\cite{2020MIE}: a multi-label classification model that obtains a term-specific representation and a status-specific representation for each utterance through an attention mechanism to determine whether the term-status pair belongs to the dialogue. 
MIE also models the multi-turn interaction simultaneously. 

\noindent\textbf{Transformer}: an encoder-decoder based generative method that treats the medical dialogue as the input and outputs a sequence containing the terms and their status. 

\noindent\textbf{MGT}~\cite{li2021medical}: a multi-granularity transformer model that models MD-TSPE similar to Transformer. 
However, unlike Transformer, MGT can simultaneously capture the role-enhanced interaction across turns and integrate mixed granularity representations to model the dialogue context fully. 

\noindent\textbf{CGT}~\cite{ye2021contrastive}: a generative model based on UniLM~\cite{DBLP:conf/nips/00040WWLWGZH19} that 
combines Transformer with contrastive learning 
to generate medical terms and their status in one stage. 

\noindent\textbf{T5}: a generative method based on pre-trained T5~\cite{raffel2020exploring}, the same backbone as our KTGF, but it generates all outputs in one stage, including medical terms and their status.

\subsection{Implementation Details}
\label{sec:Experiments Details}

We obtain experimental results of MIE based on the released codes. 
For the baseline models, including SAT, BERT-SAT, Transformer, and MGT, 
we adopt the same settings as~\cite{li2021medical} and cite the results of these models from the paper. 
In these models, the encoder of Transformer and MGT are all pre-trained with masked language modelling objective (MLM)~\cite{devlin-etal-2019-bert} on the mix of Chunyu and CMDD datasets. 
Considering that CGT does not release the source code, we only report the full results on the Chunyu dataset cited from the original paper~\cite{ye2021contrastive}. 

We report the results of T5 and KTGF with two backbones: the Chinese T5-small and T5-base~\cite{zhao2019uer}. 
The models are the base versions by default if not explicitly specified.
We implemented T5 and KTGF based on HuggingFace’s Transformers library~\cite{wolf2019huggingface} and conducted experiments using an Nvidia RTX3090 GPU. 
AdamW~\cite{loshchilov2017decoupled} is employed as the optimizer, and the weight decay is set to $0.01$.
We set the initial learning rate to $2e-5$, and the warm up step is $1000$. 
The batch size is $32$, and the total number of epochs are $100$ and $300$ for KTGF and T5, respectively.
We adopt the greedy search for generation and select the checkpoints performing best on the valid set for test evaluation. 
In the low-resource setting, to augment the training data in the term generation phase, we collect two annotated corpora that only contain term annotation, including MSL~\cite{2020Understanding} and MedDG~\cite{liu2020meddg}. 
In total, there are over $0.24$M examples.

\subsection{Full Data Evaluation}

\begin{table}[htbp]
  \centering
  \caption{Window-level evaluation results on the Chunyu and CMDD datasets.}
  \scalebox{0.8}{
  \setlength{\tabcolsep}{1mm}{
    \begin{tabular}{r|c|c|c|c|c|c|cccccc}
    \toprule
    \multicolumn{1}{c|}{\multirow{3}[6]{*}{Model}} & \multicolumn{6}{c|}{Chunyu}                      & \multicolumn{6}{c}{CMDD} \\
\cmidrule{2-13}          & \multicolumn{3}{c|}{Term} & \multicolumn{3}{c|}{Full} & \multicolumn{3}{c|}{Term} & \multicolumn{3}{c}{Full} \\
\cmidrule{2-13}          & P     & R     & F1    & P     & R     & F1    & \multicolumn{1}{c|}{P} & \multicolumn{1}{c|}{R} & \multicolumn{1}{c|}{F1} & \multicolumn{1}{c|}{P} & \multicolumn{1}{c|}{R} & F1 \\
    \midrule
    \multicolumn{1}{c|}{MIE} & 90.36 & 87.58 & 80.7  & 70.32 & 66.47 &  66.81 & \multicolumn{1}{c|}{90.85} & \multicolumn{1}{c|}{86.48} & \multicolumn{1}{c|}{87.71} & \multicolumn{1}{c|}{81.76} & \multicolumn{1}{c|}{73.7} & 79.58 \\
    \multicolumn{1}{c|}{SAT} & -  & -     & -     & - & -     & -     & \multicolumn{1}{c|}{88.5} & \multicolumn{1}{c|}{87.8} & \multicolumn{1}{c|}{87.2} & \multicolumn{1}{c|}{75.1} & \multicolumn{1}{c|}{75.3} & 74.2 \\
    \multicolumn{1}{c|}{BERT-SAT} & -  & -     & -     & - & -     & -     & \multicolumn{1}{c|}{90.7} & \multicolumn{1}{c|}{89.6} & \multicolumn{1}{c|}{89.2} & \multicolumn{1}{c|}{75.2} & \multicolumn{1}{c|}{75.2} & 74.2 \\
    \multicolumn{1}{c|}{Transformer} & 92.2  & 90.3  & 90.3  & 69.3  & 68.2  & 68    & \multicolumn{1}{c|}{90.1} & \multicolumn{1}{c|}{88.5} & \multicolumn{1}{c|}{88.7} & \multicolumn{1}{c|}{74.3} & \multicolumn{1}{c|}{72} & 72.4 \\
    \multicolumn{1}{c|}{MGT} & 93.8  & 88.6  & 90.1  & 75.3  & 71.7  & 72.7  & \multicolumn{1}{c|}{92.8} & \multicolumn{1}{c|}{90.2} & \multicolumn{1}{c|}{90.7} & \multicolumn{1}{c|}{80} & \multicolumn{1}{c|}{76.8} & 77.6 \\
    \multicolumn{1}{c|}{CGT} & -  & -     & -     & 80.53 & 78.83 & 79.42 & \multicolumn{1}{c|}{-} & \multicolumn{1}{c|}{-} & \multicolumn{1}{c|}{-} & \multicolumn{1}{c|}{-} & \multicolumn{1}{c|}{-} & - \\
    \multicolumn{1}{c|}{T5 small} & 93.15 & 91.73 & 91.5  & 74.77 & 74.28 & 73.78 & \multicolumn{1}{c|}{93.02} & \multicolumn{1}{c|}{94.34} & \multicolumn{1}{c|}{93.21} & \multicolumn{1}{c|}{83.28} & \multicolumn{1}{c|}{84.34} & 83.41 \\
    \multicolumn{1}{c|}{T5 base} & 92.19 & 92.64 & 91.49 & 75.49 & 76.26 & 75.11 & \multicolumn{1}{c|}{93.28} & \multicolumn{1}{c|}{94.28} & \multicolumn{1}{c|}{93.31} & \multicolumn{1}{c|}{83.58} & \multicolumn{1}{c|}{84.36} & 83.57 \\
    \midrule
    \multicolumn{1}{c|}{\textbf{KTGF small}} & 99.07 & 95.15 & 96.65 & 77.42 & 74.53 & 75.63 & \multicolumn{1}{c|}{\textbf{99.32}} & \multicolumn{1}{c|}{\textbf{98.16}} & \multicolumn{1}{c|}{\textbf{98.59}} & \multicolumn{1}{c|}{88.35} & \multicolumn{1}{c|}{87.4} & 87.75 \\
    \multicolumn{1}{c|}{\textbf{KTGF base}} & 99.11 & \textbf{95.72} & \textbf{97.05} & \textbf{84.31} & \textbf{81.44} & \textbf{82.55} & \multicolumn{1}{c|}{98.95} & \multicolumn{1}{c|}{97.47} & \multicolumn{1}{c|}{98.01} & \multicolumn{1}{c|}{\textbf{88.85}} & \multicolumn{1}{c|}{\textbf{87.58}} & \textbf{88.05} \\
    \textbf{w/o status} & \textbf{99.19} & 95.2  & 96.75 & 80.94 & 77.85 & 79.05 & \multicolumn{1}{c|}{99.03} & \multicolumn{1}{c|}{97.48} & \multicolumn{1}{c|}{98.06} & \multicolumn{1}{c|}{88.62} & \multicolumn{1}{c|}{87.33} & 87.81 \\
    \textbf{w/o category} & 99.03 & 95.57 & 96.91 & 84.18 & 81.34 & 82.44 & \multicolumn{1}{c|}{-} & \multicolumn{1}{c|}{-} & \multicolumn{1}{c|}{-} & \multicolumn{1}{c|}{-} & \multicolumn{1}{c|}{-} & - \\
    \textbf{w/o knowledge} & 99.11 & 95.25 & 96.72 & 80.68 & 77.68 & 78.82 & \multicolumn{1}{c|}{-} & \multicolumn{1}{c|}{-} & \multicolumn{1}{c|}{-} & \multicolumn{1}{c|}{-} & \multicolumn{1}{c|}{-} & - \\
    \bottomrule
    \end{tabular}}}%
  \label{tab:window-level-results}%
\end{table}%

\begin{table*}[htbp]
  \centering
  \caption{Dialogue-level evaluation results on the Chunyu and CMDD datasets.}
  \scalebox{0.72}{
  \setlength{\tabcolsep}{1mm}{
    \begin{tabular}{r|c|c|c|c|c|c|c|c|c|c|c|c}
    \toprule
    \multicolumn{1}{c|}{\multirow{3}[6]{*}{Model}} & \multicolumn{6}{c|}{Chunyu}                   & \multicolumn{6}{c}{CMDD} \\
\cmidrule{2-13}          & \multicolumn{3}{c|}{Term} & \multicolumn{3}{c|}{Full} & \multicolumn{3}{c|}{Term} & \multicolumn{3}{c}{Full} \\
\midrule          & P     & R     & F1    & P     & R     & F1    & P     & R     & F1    & P     & R     & F1 \\
    \midrule
    \multicolumn{1}{c|}{MIE} & 94.36 & 87.58 & 89.53 & 72.02 & 66.21 & 67.83 & 93.64 & 84.89 & 87.99 & 77.15 & 70.09 & 72.61 \\
    \multicolumn{1}{c|}{T5 small} & 90.1  & 93.37 & 90.78 & 72.69 & 77.71 & 74.05 & 87.73 & 96.46 & 91.16 & 71.09 & 77.77 & 73.7 \\
    \multicolumn{1}{c|}{T5 base} & 89.4  & 95.46 & 91.29 & 72.67 & 80.12 & 75.16 & 88.42 & 96.48 & 91.52 & 72.69 & 78.83 & 75.04 \\
    \midrule
    \multicolumn{1}{c|}{\textbf{KTGF small}} & 100   & 95.38 & 97.32 & 77.35 & 77.61 & 76.86 & \textbf{100} & \textbf{98.48} & \textbf{99.13} & 80.45 & 79.21 & 79.74 \\
    \multicolumn{1}{c|}{\textbf{KTGF base}} & \textbf{100} & \textbf{97.52} & \textbf{98.57} & 85.63 & \textbf{84.53} & \textbf{84.71} & 100   & 97.65 & 98.67 & \textbf{82.46} & \textbf{80.47} & \textbf{81.33} \\
    \textbf{w/o status} & 100   & 95.99 & 97.69 & 82.68 & 82.21 & 81.98 & 100   & 97.74 & 98.72 & 80.95 & 79.13 & 79.92 \\
    \textbf{w/o category} & 100   & 96.81 & 98.14 & \textbf{86.14} & 83.87 & 84.62 & -  & -     & -     & - & -     & -  \\
    \textbf{w/o knowledge} & 100   & 95.79 & 97.57 & 84.08 & 81.98 & 82.55 & -  & -     & -     & - & -     & -   \\
    \bottomrule  
    \end{tabular}}}%
\label{tab:dialog-level-results}%
\end{table*}%

Table~\ref{tab:window-level-results} and Table~\ref{tab:dialog-level-results} summarize the results for window-level and dialog-level on two experimental medical datasets, respectively, as described in Section~\ref{sec:Datasets and Metrics}.
The SAT method can not be evaluated on the Chunyu dataset because it needs token-level annotation. 
The window-level results in Table~\ref{tab:window-level-results} shows that our KTGF base achieves an F1-score of 82.55 on the Chunyu dataset and 88.05 on the CMDD dataset in the full training setting, 
which gains significant improvement for the complex medical conversation compared to existing methods. 
Specifically, 
on the Chunyu dataset, KTGF outperforms CGT, the previous best result, by a large margin of 3.13. 
On the CMDD dataset, our model outperforms the MGT model by 10.45. 
The improvement demonstrates the power of the two-stage generation with the knowledge-enhanced prompt. 
The dialogue-level evaluation results are shown in Table~\ref{tab:dialog-level-results}, further demonstrate the excellent performance of our proposed method.
On the Chunyu dataset, the dialogue-level evaluation typically outperforms the window-level evaluation, while the opposite is observed for the CMDD dataset. We attribute this difference to the distinct distributions of the two datasets. In the Chunyu dataset, each dialogue has fewer rounds (16.26 on average), and errors in extracting medical terms or the corresponding status based on the current window can be corrected by subsequent windows with more information, leading to higher performance at the dialogue-level evaluation. Conversely, in the CMDD dataset, each dialogue has more rounds (42.09 on average), and many dialogue windows do not mention any medical terms. The models' accurate predictions on these empty windows may result in an overestimation of its performance at the window-level evaluation, while the dialogue-level evaluation can avoid this situation. Considering that many baselines only provide window-level evaluation results, we compare our model with them based on the window-level evaluation results by default if not explicitly specified. 

\emph{Comparison with classification-based methods}: 
The SAT model is a classic baseline. 
The BERT-SAT model employs BERT as the encoder for better contextual representation, which improves the performance of the SAT model on term evaluation but not on full evaluation. 
This phenomenon illustrates that only incorporating general contextual representation does not assist the model in understanding the complex dialogue interaction, which is critical for status inference. 
In our KTGF model, prior information is incorporated into KTGF, including category and status candidates. 
Therefore, the KTGF model can more effectively infer the status of each term from the complex dialogue interaction. 
The MIE model also takes prior information into consideration. 
Specifically, to extract term-status pairs from medical dialogue, the MIE model incorporates term candidates and status candidates as part of the input and obtains a term-specific representation and a status-specific representation for each utterance through an attention mechanism. 
However, the MIE model performs poorly on term evaluation. 
For example, the gap in the F1-score on the full evaluation between the MIE model and MGT is $5.89$ on the Chunyu dataset, and the gap in the term evaluation between them is enlarged to $9.4$. 
It demonstrates that the term-specific representation hardly carries out the colloquial expression of medical terms, and the representation also ignores the correlation between terms. 
In our KTGF model, the terms are generated sequentially, which models the relationship between terms naturally. 

\emph{Comparison with generative methods}: 
From Table~\ref{tab:window-level-results} we can see that our KTGF model demonstrates better performance than other generative models on term evaluation and further gains a larger margin on full evaluation.
For example, using the same backbone model, the KTGF model improves the F1 score by 5.56 compared with T5 (97.05 vs. 91.49) on term evaluation and gains a larger margin by 7.44 (82.55 vs. 75.11) on full evaluation. 
The huge success is attributed to our two-stage generation framework and the knowledge-enhanced prompt. 
Based on the two-stage generation framework, 
we simplify the output sequence to only contain the terms in the first stage, 
which makes it more effective to learn the relationship among terms and gains better results on term evaluation. 
Moreover, the two-stage framework integrates prior knowledge more flexibly. 
With the proposed knowledge-enhanced prompt, the framework utilizes the category and status candidates in the second stage to better understand the status subtask and gains a larger margin on full evaluation.

\subsection{Low-Resource Evaluation}

\begin{table}[htbp]
  \centering
  \caption{Low-resource evaluation on the Chunyu dataset. The F1-scores of both term and full evaluations are shown.}
  \scalebox{0.85}{
    \begin{tabular}{r|c|c|c|c|c|c}
    \toprule
    \multicolumn{1}{c|}{\multirow{2}[4]{*}{Model}} & \multicolumn{2}{c|}{1\%} & \multicolumn{2}{c|}{5\%} & \multicolumn{2}{c}{10\%} \\
\cmidrule{2-7}          & \multicolumn{1}{l|}{Term} & \multicolumn{1}{l|}{Full} & \multicolumn{1}{l|}{Term} & \multicolumn{1}{l|}{Full} & \multicolumn{1}{l|}{Term} & \multicolumn{1}{l}{Full} \\
    \midrule
    \multicolumn{1}{c|}{MIE} & 27.74 & 19.25 & 43.74 & 32.35 & 68.28 & 51.12 \\
    \multicolumn{1}{c|}{T5 base} & 61.39 & 35.72 & 77.08 & 47.27 & 82.37 & 57.35 \\
    \midrule
    \multicolumn{1}{c|}{\textbf{KTGF base}} & 73.02 & 43.96 & 84.8  & 52.41 & 90.1  & 64.26 \\
    \textbf{w/ term data} & 78.79 & 44.96 & \textbf{88.19} & 55.72 & 91.87 & 65.7 \\
    \textbf{w/ both} & \textbf{79.27} & \textbf{45.24} & 87.84 & \textbf{56.27} & \textbf{94.87} & \textbf{67.81} \\
    \bottomrule
    \end{tabular}}%
  \label{tab:low resource}%
\end{table}%

Owing to the two-stage generation, our KTGF model can learn from the data that only have term annotation in the first phase. 
Moreover, we designed the special status ``not mentioned" to enrich the training data in the second phase. 
To investigate the generalization ability of KTGF equipped with these data augmentation methods, we evaluate it in a more challenging low-resource scenario. 
In the low-resource scenario, only a small percentage of training data is used. 
Here we select three different percentages as $1\%$, $5\%$, and $10\%$. 
Referring to the original dataset division for training, validation, and test sets based on the whole dialogue\cite{2020MIE}, we directly sample the whole dialogue. All the windows in these sampled dialogues are used as the low-resource training set. The trained model is then evaluated with the original test set.
We compare KTGF with the classification-based model MIE and the generative model T5 on the Chunyu dataset. 

The results are shown in Table \ref{tab:low resource}. 
As seen, KTGF consistently outperforms for all baseline models by a large margin. 
Notably, the performance gain of KTGF is even the most significant when $1\%$ training samples are used. 
For example, compared with the T5 model, our KTGF model improves the full F1-score by $6.91$ ($64.26$ for KTGF vs. $57.35$ for T5) when trained with $10\%$ samples and improves it by a larger margin of $8.24$ ($43.96$ for KTGF vs. $35.72$ for T5) when trained with $1\%$ samples. 
The results demonstrate that our model can model the MD-TSPE task more efficiently through two-stage generation and knowledge-enhanced prompt in an extremely low-resource setting.

We further explore the generalization ability of KTGF equipped with data augmentation. 
Specifically, 
w/ term data denotes that the KTGF base only augments the training data in the term generation phase, while w/ both means that the special status is further employed to enrich the training data in the status generation phase after term data augmentation. 
From Table \ref{tab:low resource}, we can see that the additional training data with term annotation improve the performance in all the metrics. 
This demonstrates that the two-stage generation of our KTGF can effectively leverage the data that only contain term annotation to alleviate the data sparsity problem. 
Then we enrich the training data in the status generation phase, and the performance is further boosted.
This indicates that our design of special status promotes KTGF learning how to generate the correct status according to the status candidates. 

\section{Analysis and discussion}
\label{sec:Analysis}

\subsection{Ablation study}
We integrate the prior schema information containing the categories and status candidates of the terms to prompt status generation. 
To analyse the contribution of the prior schema information, 
we evaluate some KTGF base variants where different prior information is removed, 
and the results are shown in Table~\ref{tab:window-level-results} and Table~\ref{tab:dialog-level-results}. 
Specifically, 
w/o status denotes that the KTGF base, which do not use the status information and only use the category information, while w/o category is the opposite of w/o status; w/o knowledge means that neither status nor category information is utilized. 
Since the CMDD dataset only contains the symptom category, it is unnecessary to distinguish terms from different categories. Therefore, we do not use the category information in this dataset.

From Table~\ref{tab:window-level-results}, it can be seen that each time after removing a kind of relevant prior knowledge, the performance of the KTGF model declines, indicating that all kinds of knowledge are helpful for the MD-TSPE task. 
On the Chunyu dataset, the F1-score drops from $82.55$ to $79.05$ after removing the status candidates, which is a larger reduction ($3.5$) compared with the w/o category ($0.11$). 
That is because the status candidates intuitively include the information that the generated output needs, which makes the model learn the status generation more effectively. 
The performance after removing both categories and status candidates is the worst. 
Table \ref{tab:dialog-level-results} also shows that incorporating both status and category information gives KTGF the best results on the dialogue-level evaluation.

We also note the small impact of introducing prior schema information on term generation. 
As shown in Table~\ref{tab:window-level-results}, 
both the precision on the Chunyu, all metrics on the CMDD dataset are 
improved after removing the status candidates, which is different from the full evaluation. 
The preliminary analysis is that the status candidates depend on the categories and have different effects on different categories of terms. 
For further analysis, we conduct the evaluation on different categories, and the details are left to Section \ref{sec:category}. 

In addition, compared with the results on the CMDD dataset, 
more remarkable improvement is achieved on the Chunyu dataset after introducing the prior schema information. 
One possible reason for explaining this phenomenon 
is that the Chunyu dataset has more categories and more status candidates, and the learning of complex prior task knowledge is more dependent on the knowledge-enhanced prompt.

\begin{figure*}[t]
\centerline{\includegraphics[height=8.5cm,width=14cm]{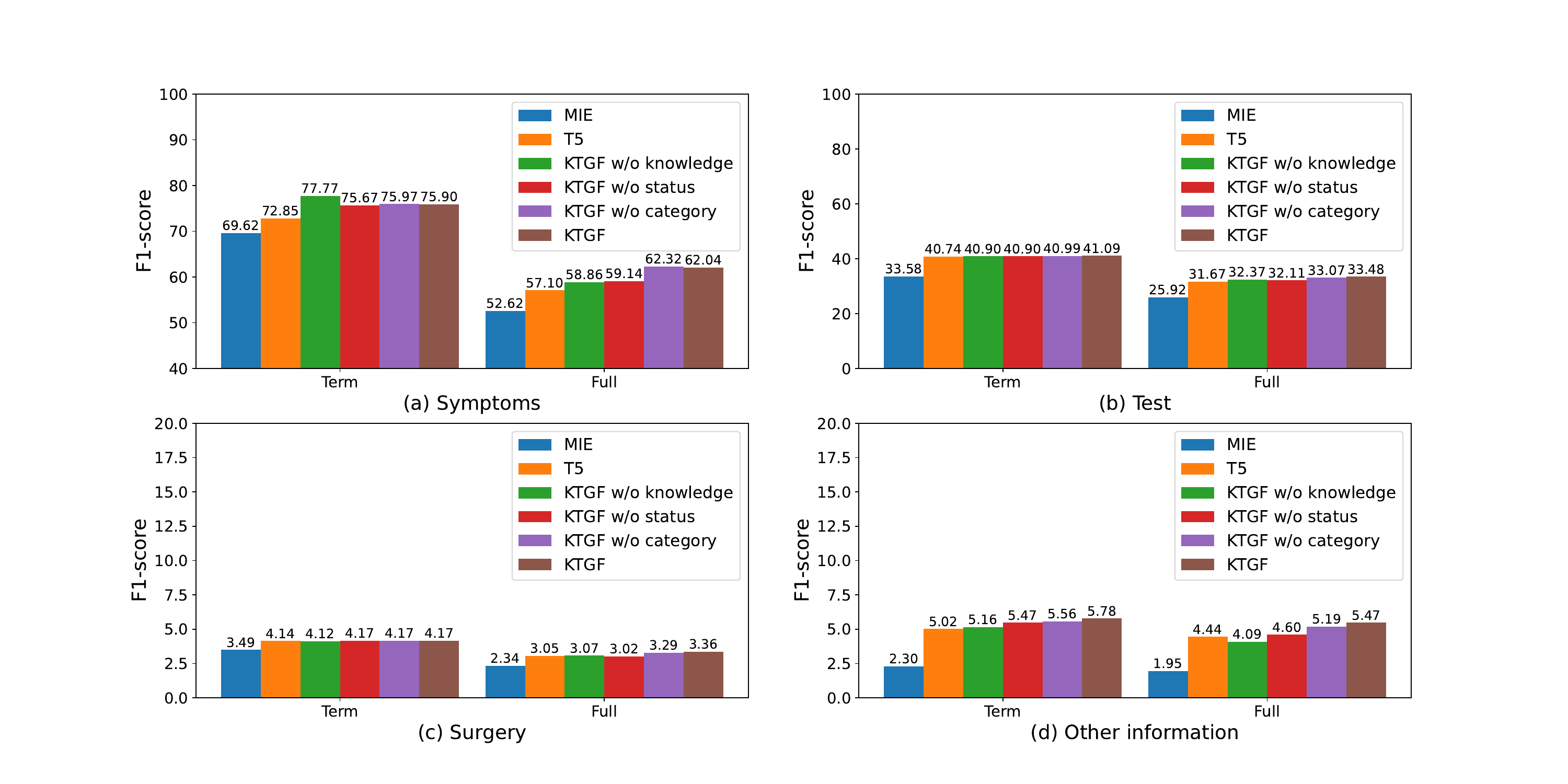}}          
\caption{The evaluation of different categories on F1-score.}
\label{category_exp}                     
\end{figure*}                    

\subsection{Comparison w.r.t Different categories}
\label{sec:category}

Considering that the prior schema information contains the category and the status candidates of terms, and the status candidates also depend on the categories,  
we are curious about the effectiveness of the knowledge on different categories.
Here we evaluate the Chunyu test set based on the categories, and the empty prediction is not taken into account. 
The Chunyu dataset has four categories (Symptom, Test, Surgery, and Other Info), and the evaluation results of different categories are shown in Figure \ref{category_exp}.

As seen, the performance of all models on different categories varies greatly, but it is basically consistent with the amount of data for corresponding categories in the dataset. The category ``Symptom" occurs most frequently, while the ``Surgery" and ``Other info" rarely appear. 
For all categories, our proposed KTGF outperforms the baseline models MIE and T5. 
After utilizing the categories and status candidates, KTGF performs best both on term evaluation and full evaluation in three of the four categories, except ``Symptom". 
Considering that ``Symptom" occurs most frequently in the dataset among all the categories, we analyse that it is the utilization of both category and status candidates that makes the model pay more attention to the long-tail category. 
Therefore, the performance of the category ``Symptom" is reduced, which results in KTGF w/o knowledge performing the best on term evaluation. 
Even so, compared with KTGF w/o knowledge, the KTGF w/o category still performs better on full evaluation, 
demonstrating that status knowledge can be helpful for the category ``Symptom" on status generation. 

\begin{figure*}[t]

\centerline{\includegraphics[height=5cm,width=14cm]{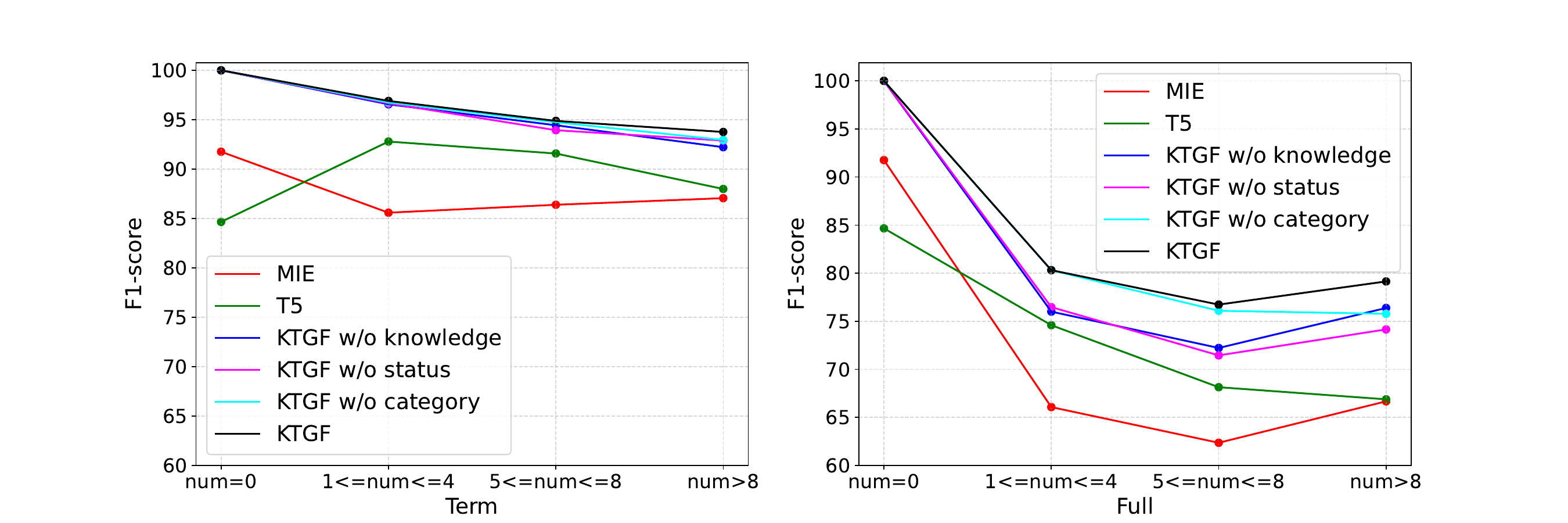}}    
\caption{The evaluation for different numbers of mentioned terms on F1-score.}
\label{term_exp}                     
\end{figure*}                    

\subsection{Comparison w.r.t The numbers of mentioned terms}
\label{sec:term}

The two-stage generation of our KTGF model simplifies the output sequence to only contain the terms in the first stage and gains better results on term evaluation. 
Here we further explore the effectiveness of the KTGF model on different numbers of mentioned terms, which determines the length and complexity of the output sequence in the term generation phase. 
All the experiments are performed on the Chunyu dataset, and the results are presented in Figure \ref{term_exp}. 
``$num=0$" means that the test examples do not have golden labels. 
``$1\le num \le 4$" means there are no more than four terms mentioned in the test examples. 

It can be observed that the performance of the classification-based model MIE has no obvious relationship with the number of mentioned terms in the term evaluation. 
With the increase in the number of mentioned terms, the performance of the T5 model decreases significantly both on term evaluation and full evaluation. 
Even the F1-score of the T5 model in "$num=0$" only is $85$, 
which means the model can not identify the examples that do not have golden terms. 
Compared with the generative model T5, the performance of the two-stage generation decreases more slowly as the number of mentioned terms increases on term evaluation. Even the downwards trend of KTGF has changed on full evaluation. 
These results show that KTGF is more robust both in the situation of no terms and many terms
during term generation and eliminates the interference of the number of
generated terms during status generation. 
Unsurprisingly, after utilizing the categories and status candidates, the KTGF model gains the best results on full evaluation.

\subsection{Comparison w.r.t The changed status during dialogue interaction}
\label{sec:status}

A common scenario is that the status of the term is changed through the interaction of speakers. 
Sometimes the status of the term may not appear immediately after the term, and status inference requires a better understanding of dialogue interaction.
We filter the examples that contain these terms from the Chunyu test set and evaluate them with different models. 
The results are shown in Table \ref{status results}. 
From the results, we can see that our KTGF model achieves the best performance for all metrics. 
Compared with the T5 model, the KTGF model even gains a margin of $9.57$ ($73.48$ for KTGF vs. $63.91$ for T5), which is more than the margin of $7.44$ obtained on overall data ($82.55$ for KTGF vs. $75.11$ for T5) for full test evaluation. 
This demonstrates that our model can better understand more complicated dialogues where the status changes as dialogue progresses. 

\begin{table}[htbp]
  \centering
  \caption{Evaluation of the examples with the changed status.}
  \scalebox{0.85}{
    \begin{tabular}{r|c|c|c|c|c|c}
    \toprule
    \multicolumn{1}{c|}{\multirow{2}[4]{*}{Model Status}} & \multicolumn{3}{c|}{Term} & \multicolumn{3}{c}{Full} \\
\cmidrule{2-7}          & Precision     & Recall     & F1-score    & Precision     & Recall     & F1-score \\
    \midrule
    \multicolumn{1}{c|}{MIE} & 93.4  & 81.23 & 85.09 & 58.85 & 59.51 & 57.43 \\
    \multicolumn{1}{c|}{T5 base} & 94.91 & 93.43 & 93.32 & 64.86 & 64.08 & 63.91 \\
    \midrule
    \multicolumn{1}{c|}{\textbf{KTGF base}} & \textbf{100} & 95.23 & 97.21 & \textbf{75.69} & \textbf{71.92} & \textbf{73.48} \\
    \textbf{w/o status} & 100   & 94.78 & 96.92 & 71.31 & 67.66 & 69.15 \\
    \textbf{w/o category} & 100   & \textbf{95.3} & \textbf{97.28} & 75.24 & 71.69 & 73.19 \\
    \textbf{w/o knowledge} & 100   & 94.98 & 97.04 & 71.62 & 67.97 & 69.46 \\
    \bottomrule
    \end{tabular}}%
  \label{status results}%
\end{table}%

\begin{table}[htbp]
  \centering
  \caption{A case inferred by baseline models and the proposed model.}
  \scalebox{0.85}{
  \resizebox{\textwidth}{100mm}{
    \begin{tabular}{lp{10cm}}
    \toprule
    \multicolumn{2}{l}{\textbf{Dialogue}} \\
    \midrule
    Patient: & 22 \\
          & \multicolumn{1}{p{21.445em}}{22 years old.} \\
    Doctor: & \begin{CJK*}{UTF8}{gbsn}做过检查没有。怎么个疼法？ \end{CJK*}\\
          & Have you done any inspections? What's the details of the pain? \\
    Patient: & \begin{CJK*}{UTF8}{gbsn}去年这个时候我也会有这种情况。但是做了很多检查都与心脏无关。是吃心可舒好转的。按理说我的疼应该 和心脏没关系。具体疼痛我也不知道从哪里出来的。你深呼吸，左胸会疼。就这3天。会不会是因为天气变化\end{CJK*} \\
          & It also appeared last year. But many tests have shown that it has nothing wrong with the heart. It takes a turn for the better after taking Xinkeshu Capsule. My pain should have nothing to do with my heart. I don't know where the pain comes from. My left chest would pain when I toke a deep breath. It appears just for three days. Could it be from the changed weather? \\
    Doctor: & \begin{CJK*}{UTF8}{gbsn}感冒了么。心电图，心肌酶做了吗。抽烟么 \end{CJK*}\\
          & Have you caught a cold? Have you done electrocardiogram and myocardial enzyme? Do you smoke ? \\
    Patient: & \begin{CJK*}{UTF8}{gbsn}没有\end{CJK*} \\
          & None. \\
    \midrule
    \multicolumn{2}{l}{\textbf{Term-Status Pairs Label}} \\
    \midrule
    \multicolumn{2}{l}{\begin{CJK*}{UTF8}{gbsn}胸痛: 阳性\end{CJK*}(Chest pain: appear)} \\
    \multicolumn{2}{l}{\begin{CJK*}{UTF8}{gbsn}心电图: 患者未做\end{CJK*}(Electrocardiogram: not done)} \\
    \multicolumn{2}{l}{\begin{CJK*}{UTF8}{gbsn}感冒: 阴性\end{CJK*}(Cold: absent)} \\
    \multicolumn{2}{l}{\begin{CJK*}{UTF8}{gbsn}吸烟: 异常\end{CJK*}(Smoking: abnormal)} \\
    \multicolumn{2}{l}{\begin{CJK*}{UTF8}{gbsn}心肌酶: 患者未做\end{CJK*}(Myocardial enzyme: not done)} \\
    \midrule
    \multicolumn{2}{l}{\textbf{MIE}} \\
    \midrule
    \multicolumn{2}{l}{\begin{CJK*}{UTF8}{gbsn}胸痛: 阳性\end{CJK*}(Chest pain: appear)} \\
    \multicolumn{2}{l}{\begin{CJK*}{UTF8}{gbsn}心电图:患者未做\end{CJK*}(Electrocardiogram: not done)} \\
    \multicolumn{2}{l}{\begin{CJK*}{UTF8}{gbsn}感冒: 未知\end{CJK*}(Cold: unknown)} \\
    \multicolumn{2}{l}{\begin{CJK*}{UTF8}{gbsn}心肌酶: 患者未做\end{CJK*}(Myocardial enzyme: not done)} \\
    \multicolumn{2}{l}{\begin{CJK*}{UTF8}{gbsn}心肌酶: 未知\end{CJK*}(Myocardial enzyme: unknown)} \\
    \midrule
    \multicolumn{2}{l}{\textbf{T5 base}} \\
    \midrule
    \multicolumn{2}{l}{\begin{CJK*}{UTF8}{gbsn}胸痛: 阳性\end{CJK*}(Chest pain: appear)} \\
    \multicolumn{2}{l}{\begin{CJK*}{UTF8}{gbsn}心电图:患者未做\end{CJK*}(Electrocardiogram: not done)} \\
    \multicolumn{2}{l}{\begin{CJK*}{UTF8}{gbsn}感冒: 阴性\end{CJK*}(Cold: absent)} \\
    \multicolumn{2}{l}{\begin{CJK*}{UTF8}{gbsn}心肌酶: 患者已做\end{CJK*}(Myocardial enzyme: done)} \\
    \multicolumn{2}{l}{\begin{CJK*}{UTF8}{gbsn}呼吸困难: 阳性\end{CJK*}(Dyspnea: appear)} \\
    \midrule
    \multicolumn{2}{l}{\textbf{KTGF base}} \\
    \midrule
    \multicolumn{2}{l}{\begin{CJK*}{UTF8}{gbsn}胸痛: 阳性\end{CJK*}(Chest pain: appear)} \\
    \multicolumn{2}{l}{\begin{CJK*}{UTF8}{gbsn}心电图:患者未做\end{CJK*}(Electrocardiogram: not done)} \\
    \multicolumn{2}{l}{\begin{CJK*}{UTF8}{gbsn}感冒: 病人阴性\end{CJK*}(Cold: absent)} \\
    \multicolumn{2}{l}{\begin{CJK*}{UTF8}{gbsn}吸烟: 异常\end{CJK*}(Smoking: abnormal)} \\
    \multicolumn{2}{l}{\begin{CJK*}{UTF8}{gbsn}心肌酶: 患者未做\end{CJK*}(Myocardial enzyme: not done)} \\
    \bottomrule
    \end{tabular}}}%
  \label{case study}%

\end{table}%

\subsection{Case Study}
To more qualitatively compare the performance of the KTGF model with other baseline models, we select an example from the test set of Chunyu, as shown in Table \ref{case study}. 
There are five terms mentioned in this example, 
\texttt{\small electrocardiogram} and \texttt{\small myocardial enzyme} belong to the test, the category of \texttt{\small chest pain} and \texttt{\small cold} is the symptom, and \texttt{\small smoking} belongs to other information. 

As seen, all baselines ignore 
\texttt{\small smoking}, while our KTGF model produces it and its corresponding status correctly. 
This demonstrates the satisfactory ability of KTGF to capture the terms of rare categories. 
In addition, the T5 model misjudges the breath mentioned in the spoken description of chest pain as \texttt{\small dyspnea}, which is not the case in the KTGF model. 
It shows that the KTGF model can better understand the colloquial expression in medical dialogues. 
Moreover, the last sentence changes the statuses of many terms, and only our model predicts all of them correctly, which verifies the effectiveness of our KTGF model in complex dialogue interaction scenarios.

\section{Conclusion}
\label{sec:Conclusion}

This paper proposes a knowledge-enhanced two-stage generative framework (KTGF) for term-status pair extraction from medical dialogues (MD-TSPE). 
In the framework, we generate all terms first and then generate the status of each generated term in the second phase. 
We further design the knowledge-enhanced prompt in the second phase to leverage the category and status candidates of the generated term. 
For the low-resource setting, we design a special status ``not mentioned", which makes more terms available and enriches the training data in the second phase.

We evaluated our KTGF model on two Chinese medical dialogue datasets, Chunyu and CMDD. The experiments show that KTGF surpasses the previous state-of-the-art results by 5.55 and 5.28 for Chunyu, and 3.13 and 4.48  for CMDD on term and full evaluations, respectively. It demonstrates the advantages of both two-stage generation and knowledge-enhanced prompt 
in complex dialogue scenarios compared with existing classification-based models and generative models. 
Low-resource experiments show that two-stage generation can leverage data with only term annotation, and improve the term generation performance. Our designed special status ``not mentioned" can further enhance status generation. 
The effectiveness of different components of the proposed framework is also illustrated by the ablation study and further discussion.
Based on the analysis in this paper, we hope that more works can be inspired to complete better MD-TSPE as well as other similar tasks through step-by-step generation and find ways to integrate richer prior knowledge into the model.

\section*{Declaration of competing interests}

The authors declare that they have no known competing financial interests or personal relationships that could have appeared to influence the work reported in this paper.

\section*{Acknowledgments}
This work is supported by the Key Research Program of the Chinese Academy of Sciences under Grant (No.ZDBS-SSW-JSC006) and the National Natural Science Foundation of China (No.62206294).

\bibliographystyle{bst/sn-mathphys}
\bibliography{sn-bibliography}

\end{document}